\documentclass{article}

\usepackage{spconf,amsmath,graphicx}
\usepackage{amsfonts}
\usepackage{hyperref}
\usepackage{xcolor}
\def\x{{\boldsymbol{x}}}

\title{Greenery Segmentation In Urban Images By Deep Learning}
\name{Artur Andr\'e A.M. Oliveira, Nina S. T. Hirata, R. Hirata Jr.
\thanks{We were sponsored by the FAPESP, grant \# 2018/10767-0, S\~ao Paulo Research Foundation (FAPESP), and acknowledge Microsoft for a sponsorship through the Microsoft Azure Sponsorship.}}
\address{Universidade de S\~ao Paulo\\
	Instituto de Matem\'atica e Estat\'istica}
\begin{document}
%\ninept
%
\maketitle
\begin{abstract}
Vegetation is a relevant feature in the urban scenery and its awareness 
can be measured in an image by the Green View Index (GVI). 
Previous approaches to estimate the GVI
were based upon heuristics image processing approaches and recently
by deep learning networks (DLN). By leveraging some recent DLN architectures 
tuned to the image segmentation problem and exploiting a weighting strategy
in the loss function (LF) we improved previously reported results in similar 
datasets.
\end{abstract}
\section{Introduction}
\label{sec:intro}
Vegetation is an important feature in a urban scenery because it helps 
dissipation of heat-islands, improves air quality, saves energy by cooling 
the ambient etc~\cite{lovasi2008children, mcpherson1997quantifying, nowak2014tree}. 
Besides that, greenery also plays an important role in health and landscape 
aesthetics~\cite{kardan2015neighborhood, lothian1999landscape, thayer1978plants}. 

Automatic greenery and general vegetation segmentation in urban images dates back 
to 2009 when Yang et al.~\cite{yang2009can} developed the Green View Index (GVI), 
a measure to quantify the amount of visible urban forestry, validated 
by correlating the amount of greenery from a street-level image and the canopy 
cover measured from satellite imagery. 

Later, Li et al.~\cite{li2015assessing} launched the Treepedia project and a better estimation of the GVI based on Mean Shift image segmentation technique to Google Street View (GSV) images. In 2018, Li et al.~\cite{cai2018treepedia} released Treepedia 2 where the segmentation method is a modified version of a DLN architecture, the PSPNet~\cite{DBLP:journals/corr/ZhaoSQWJ16}. 
They mainly proposed a reduction from 19 sigmoid units from the last layer, originally used to classify a pixel as one of 19 classes from Cityscapes dataset~\cite{Cordts2016Cityscapes}, to just two to classify a pixel as greenery or non-greenery. Besides that, they proposed a modified version of the ResNet architecture~\cite{DBLP:journals/corr/HeZRS15} to estimate the GVI of an image directly.

Motivated by the DLN success to perform image segmentation, the so called semantic segmentation~\cite{DBLP:journals/corr/PohlenHML16, DBLP:journals/corr/ChenPK0Y16, DBLP:journals/corr/LinMS016, Yang_2018_CVPR, DBLP:journals/corr/abs-1802-02611, valada2017adapnet}, 
we compared the performance of multiple architectures to estimate the amount of greenery in an image. We propose a weighting strategy based on the frequency of the classification classes to improve previous results. Tests has been done over two datasets, the CamVid~\cite{fauqueur2007assisted} dataset and an in-house~\cite{github:groundtruth} built for this application. 

\section{DLN architectures for image segmentation}
In this section we review state of the art architectures for semantic segmentation used in the experiments. The basic idea of these architectures is that the first 
part of the network extract features and down samples the image while the second part mix pixels with similar characteristics and up samples the image resulting in a segmentation.

AdapNet~\cite{valada2017adapnet} has as a contracting segment followed by an expansive one, with residual blocks allowing a greater number of layers and multi-resolution blocks, the residual blocks. These blocks are composed by atrous convolutional filters (ACF) with distinct dilation rates that are combined by concatenating their outputs.
DenseASPP~\cite{Yang_2018_CVPR} combines multiple ACFs with increasing dilation rates in a cascaded way by combining the input layer and features from earlier ACFs (with 
smaller dilation rates) in order to have more features for more scales than those obtained in~\cite{DBLP:journals/corr/ChenPK0Y16}. 
DeepLabV3+ explores the multi-resolution aspect of semantic segmentation by exploiting features from atrous spatial pyramid pooling (ASPP)~\cite{DBLP:journals/corr/ChenPK0Y16} and the encoder-decoder architecture.
It encodes the input with an ASPP scheme and, instead of up sampling directly to the 
final resolution, it uses a decoder scheme with Convolutional Layers (CL) after concatenating the feature map produced before the ASPP encoding with the up sampled result from the ASPP encoding. 
SegNet~\cite{badrinarayanan2015segnet} has a VGG style~\cite{DBLP:journals/corr/SimonyanZ14a} 
encoder followed by a corresponding decoder. The contribution lies on its up sampling scheme where a convolution is done over an activation map containing only values from indices chosen from a directly previous max-pooling layer. 
Encoder-Decoder-Skip (EDS)~\cite{semsegsuiten} is a variation of SegNet where the up sampling is done by transposed convolutions and the skip connections are made from each layer at the encoder to the corresponding layer at the decoder part. 
In Full-Resolution Residual Networks (FRRN)~\cite{DBLP:journals/corr/PohlenHML16}, multi-resolution features are obtained through the combination of two paths, one with convolutions and residual connections, to keep low level features used to extract localization information, and the other composed by convolutions 
and pooling operations that are combined using a Full-Resolution Residual Unit (FRRU). A FRRU have an input from each path and one output back to each path. This allows different multiple resolution features to be selectively used via weights learned at the Full-Resolution Residual Units.
%
%In this work both implementations of FRRN (A and B) are tested, the main difference between FRRN-A and FRRN-B 
%is that the B version has one more pooling/up sampling pair. 
%
The UNet architecture~\cite{DBLP:journals/corr/RonnebergerFB15} is composed by a encoder-decoder like architecture
%contracting path (encoder), composed by convolutional and max-pooling layers, and a expanding path (decoder), composed by 
with the skip connections between each pair of pooling/up sampling layers done by concatenating the output from a pooling with the same resolution up sampled feature map. 
The main strength of this architecture comes from a custom LF with a exponential weighting for boundaries, i.e., samples with classes very less frequent than others in the dataset. 
%
%We explore a similar idea in this paper as presented in Section~\ref{sec:methodology}. 
%
The MobileUNet-Skip~\cite{semsegsuiten} is a variation of the UNet architecture in which convolutions from the encoder path are done with separable convolutional layers~\cite{DBLP:journals/corr/HowardZCKWWAA17}. 
%In other words, a layer that computes an intermediate feature map with as many channels as the number of channels of its input feature map multiplied by some positive integer and produces its output by applying a point-wise convolution (1x1 convolution) with as many channels as required, over the intermediate layer, thus allowing a depth-wise separable convolution. The decoding path is up sampled using transposed convolutions.
In the PSPNet~\cite{DBLP:journals/corr/ZhaoSQWJ16}, features are firstly extracted 
by CLs and then processed by the Pyramid Pooling Module, composed by an arbitrary number of independent pooling layers, having different sizes and thus producing feature maps with different sizes. Those feature maps are up sampled directly to the original size before the pooling operation and are concatenated together with the 
original feature map. That concatenated feature map is then fed to another CL 
where the final prediction will be generated.

\section{Methodology}
\label{sec:methodology}

To estimate the GVI from an image, we formulate the problem as a segmentation one, 
so that the GVI of an image will be the proportion of that image classified as vegetation. This 
approach can be generalized to a set of images as done in~\cite{treepedia} summing the amount of 
greenery in each image and dividing it by the sum of the areas of each image. 

The segmentation is based on DLN, where multiple architectures were trained using the 
CamVid dataset and tested on both CamVid and INACITY's Greenery dataset, an in-house
dataset created for this application that can be accessed (as well as code used to produce our results) at~\cite{github:groundtruth}.

CamVid dataset is composed of video sequences and to the best of our knowledge 
only one is fully annotated, with 701 labeled images~\cite{fauqueur2007assisted}. Those annotations comprehend 32 classes, including a class "void" (pixels that are ambiguous with relation to which class they belong to). For weighting and performance
computations, we treat the void pixels as missing data, and as such, they are disregarded. For experimental purposes, the dataset is split into 421 images for 
training, 112 for validation and 168 for testing. 

INACITY's Greenery dataset contains 100 images manually annotated and classified in three classes (greenery and non-greenery), including the "void" class. The procedure taken to produce INACITY's greenery dataset and CamVid's dataset differs to the purpose each one was created. Because the latter is for autonomous vehicles training, an extra care is taken with respect to roads, signs and its boundaries segmentation while the former was designed to be precise in respect to the greenery, so that an extra care was taken to not include pixels around leaves that are not greenery (e.g. sky pixels that appears among the sparse trees' canopies).

We trained a EDS network using weighted cross entropy LF as shown in 
Eq.~\ref{eq:weighted_loss} where a labeled sample, that is, the expected prediction for a given pixel in an image is encoded using the one-hot encoding (a binary array of matrices representation where the label of a pixel is the value of the pixel in the matrix where the array is not zero)~\cite{harris2010digital}. From now on we will refer to the expected prediction as \textit{matrix of labels}.

The LF will therefore be the a weighting of the cross-entropy between the 
soft-max of each position of the activation matrix of the last layer of the network 
and the expected vector. Borrowing the notation used in~\cite{ronneberger2015u}, the 
soft-max function is defined as in Eq.~\ref{eq:softmax}.
\begin{equation}
\begin{split}
    p_c(\x) = \frac{exp(a_c(\x))}{\sum_{c'=1}^{C} exp(a_{c'}(\x))} \\
    p(\x) = [p_1(\x), ..., p_C(\x)]
    \label{eq:softmax}
\end{split}
\end{equation}

Where:
\begin{itemize}
    \item C is the number of classes
    \item c is the channel corresponding to a specific class of the dataset
    \item $a_c(\x)$ is value for the position $\x \in \Omega \subset \mathbb{Z^2}$ of the activation matrix at the $c \in {1, ..., C}$ channel
\end{itemize}
\noindent{The weighting function is computed over each image of the dataset as in Eq.~\ref{eq:weighting_func}.}
\begin{equation}
\begin{split}
    \omega_{c}^{C}(y) = \frac{|y|}{|y_c|*C}\\
    \omega(y) = [\omega_{1}^{C}, ..., \omega_{C}^{C}]
    \label{eq:weighting_func}
\end{split}
\end{equation}
\noindent{Where:}
\begin{itemize}
    \item $|y|$ is the number of positions considered in $y$ (e.g. the number of pixels not labeled as "void" in an image)
    \item $|y_c|$ is the number of positions assigned to class $c$
    \item $C$ is the number of classes on the dataset
\end{itemize}
\noindent{The "void" is not counted in the summation.}

Once we have the loss for each pixel and its corresponding weight vector, we define the weighted loss 
function for a given image $i$ as in Eq.\ref{eq:weighted_loss}:
\begin{equation}
L_i(y, \hat{y}) = - \sum_{\x \in \Omega} (y(\x) * log(\hat{y}(\x)) * \omega(y)^\top
\label{eq:weighted_loss}
\end{equation}
\noindent{Where:}
\begin{itemize}
    \item $y(\boldsymbol{x})$ is the expected labels tensor
    \item $\hat{y}(\boldsymbol{x}) = p(\boldsymbol{x})$ is the predicted labels matrix
    \item $\omega(y)$ is the weighting vector calculated in equation \ref{eq:weighting_func}.
\end{itemize}
\noindent{In order to weight the LF, first we compute the sum of the element-wise 
multiplication between the logarithm of each prediction vector and its corresponding expected prediction vector, resulting in a vector with the accumulated error for each class. Then, another element-wise multiplication with a vector of weights as computed in Eq.~\ref{eq:weighting_func}. This is similar to the custom loss used in~\cite{DBLP:journals/corr/RonnebergerFB15} where an exponential function is used to increase the weight of cells' boundaries biomedical images. This approach is   important when one faces too much imbalance between classes. When it comes to classifications with only two classes (greenery and non-greenery), the approach prevents the network from classifying all pixels as non-greenery.}

\section{Experiments}
\label{sec:experiments}
In this section we present some of the experiments to assess the greenery of two different datasets using different DLN architectures and the training and validation protocols explained in Section~\ref{sec:methodology}.  

We first trained all the considered architectures: AdapNet~\cite{valada2017adapnet}, 
DeepLabV3+~\cite{DBLP:journals/corr/abs-1802-02611}, DenseASPP~\cite{Yang_2018_CVPR}, 
Encoder-Decoder-Skip (EDS)~\cite{semsegsuiten}, Full-Resolution Residual Networks 
(FRRN-A and B)~\cite{DBLP:journals/corr/PohlenHML16}, MobileUNet-Skip~\cite{DBLP:journals/corr/HowardZCKWWAA17} and RefineNet~\cite{DBLP:journals/corr/LinMS016}, using the CamVid dataset.
The Skip suffix for MobileUNet and for Encoder-Decoder architectures means that there is a residual connection (also known as skip-connection) between each layer of the encoding path (down sampling) and its corresponding layer at the decoding path (up sampling). Note that at this first test neither the networks nor the datasets were adjusted, that is, no weighting over the LF, nor dataset normalization operations were performed.

We tested the architectures on the same dataset and evaluated their mean Intersection over Union score (IoU), also known as Jacquard index, as well as the average run time (in milliseconds) per image (see Table~\ref{tab:pre_comparison}). The performance of the architectures for IoU are usually better than $50\%$ and the best ones are the EDS and the FRRN-A and B. In terms of run time, The FRRNs are the worst and EDS has a good compromise between performance and run time.  

\begin{figure*}[!ht]
    \centering
    \begin{minipage}[t]{0.24\textwidth}
        \includegraphics[width=\linewidth]{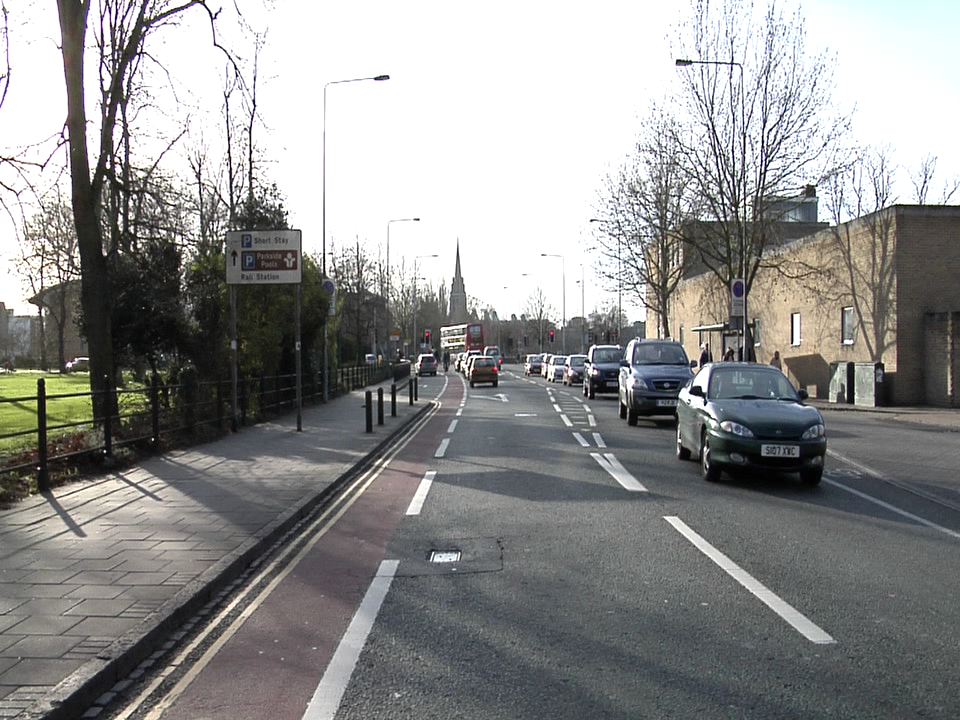}
        \caption{Image 0016E5\_02310 from CamVid's dataset}
        \label{img:camvid_original}
    \end{minipage}\hfill
    \begin{minipage}[t]{0.24\textwidth}
        \includegraphics[width=\linewidth]{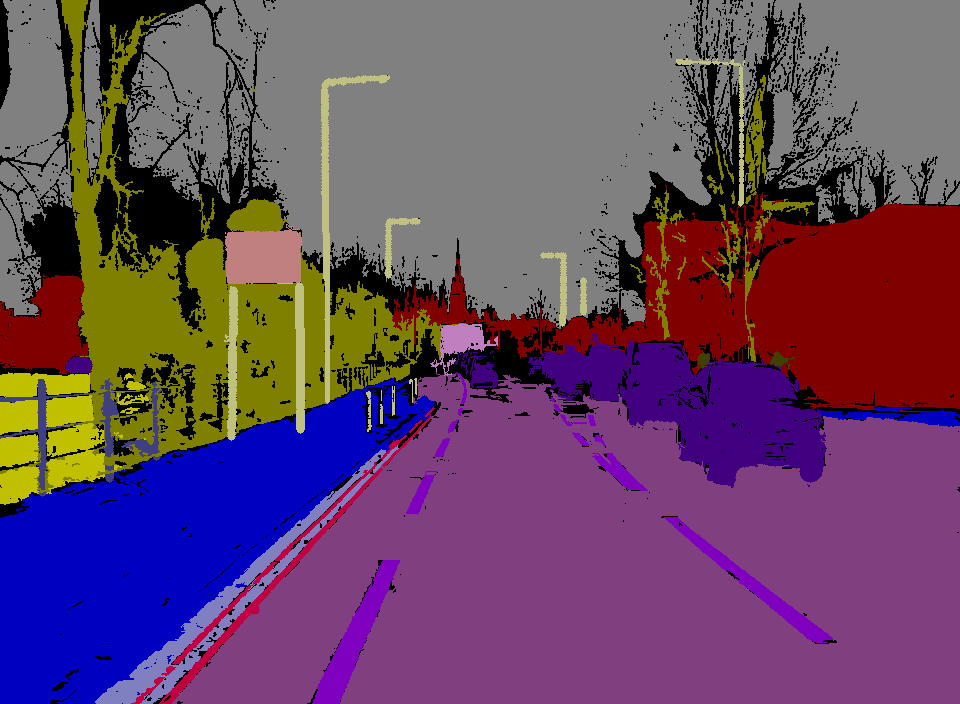}
        \caption{GT 32 classes + void}
        \label{img:camvid_GT}
    \end{minipage}\hfill    \begin{minipage}[t]{0.24\textwidth}
        \includegraphics[width=\linewidth]{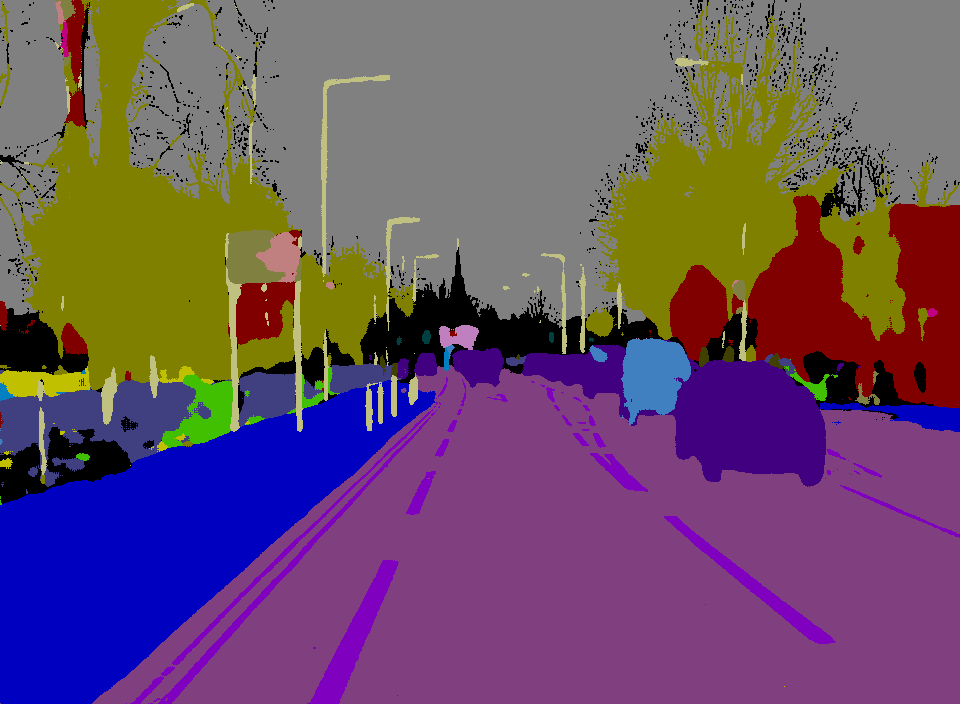}
        \caption{32 classes no weighting}
        \label{img:camvid_no_weighted_pred}
    \end{minipage}\hfill
    \begin{minipage}[t]{0.24\textwidth}
        \includegraphics[width=\linewidth]{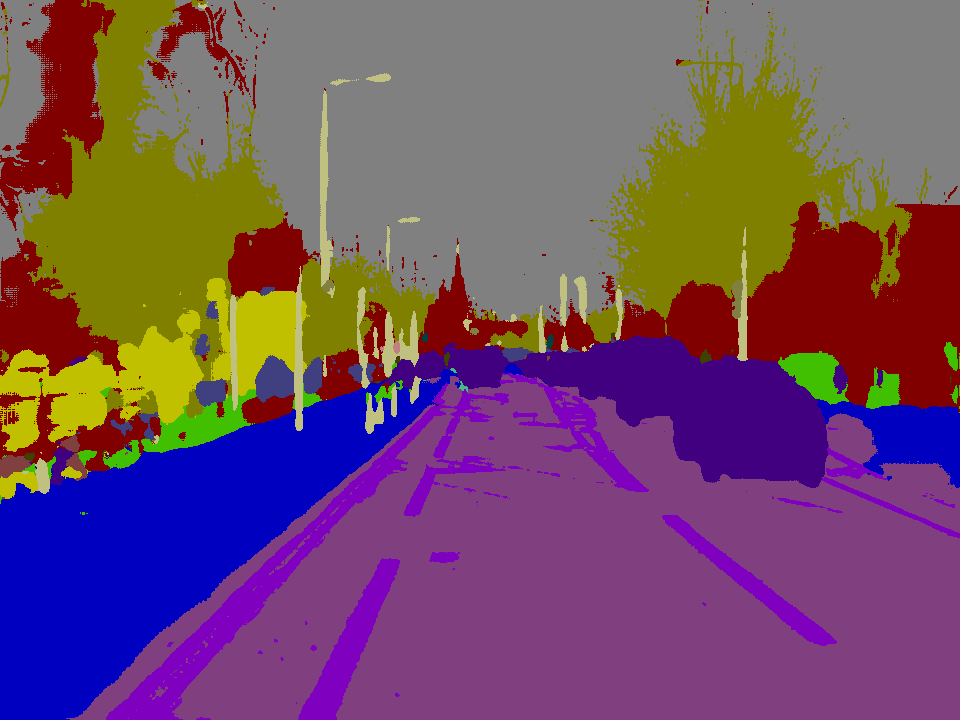}
        \caption{32 classes with weighting}
        \label{img:camvid_weighted_pred}
    \end{minipage}
\end{figure*}

We also tested the architectures trained over CamVid (with no adjustments) in INACITY's Greenery dataset~\cite{github:groundtruth} to assess the robustness of the architectures (see Table~\ref{tab:pre_greenery}). The results are all worst than $50\%$ as expected because the dataset as explained before have different purposes. 

\begin{table}[ht]
\centering
\caption{Architectures and their IoU score and average running time per image with 32 classes in the dataset (no-weighting).}
\begin{tabular}{|l|c|c|}
\hline
Architecture         & IoU & Run time \\
\hline
AdapNet              & 20\%  & 184 ms  \\
DeepLabV3+           & 49\%  & 257 ms  \\
DenseASPP            & 37\%  & 107 ms  \\
EDS & 59\%  & 433 ms  \\
FRRN-A               & 59\%  & 779 ms  \\
FRRN-B               & 60\%  & 798 ms  \\
MobileUNet-skip      & 54\%  & 332 ms  \\
PSPNet               & 51\%  & 220 ms   \\
RefineNet            & 54\%  & 395 ms   \\
\hline
\end{tabular}
\label{tab:pre_comparison}
\end{table}

\begin{table}[ht]
\centering
\caption{Architectures and their IoU score and average running time per image for INACITY greenery dataset (no-weighting).}
\begin{tabular}{|l|c|c|}
\hline
Architecture         & IoU  & Run time \\
\hline
AdapNet              & 36\% & 293 ms   \\
DeepLabV3+           & 37\% & 205 ms   \\
DenseASPP            & 37\% & 107 ms   \\
EDS & 35\% & 742 ms   \\
FRRN-A               & 42\% & 1285 ms   \\
FRRN-B               & 45\% & 1238 ms   \\
MobileUNet           & 38\% & 541 ms   \\
PSPNet               & 42\% & 390 ms   \\
RefineNet            & 46\% & 588 ms   \\
\hline
\end{tabular}
\label{tab:pre_greenery}
\end{table}

To assess the weighting strategy, we choose EDS for the reasons stated before and  trained it using the CamVid dataset and applied the method described in Section~\ref{sec:methodology}. We also tested a classes reduction strategy and trained with less classes. The rational is to test the effect of having less classes by keeping only 6 classes and relabelling the others as a custom class named "others". The classes used for training are: Road, Road Shoulder, Sidewalk, Sky, Tree, Vegetation Misc. and "Others". The "void" label was not included in "others". 

We tested the new approaches the same way as before and the results are shown in Table~\ref{tab:enc_camvid_greenery}. The first column of the table shows the number of classes used during training, the second column shows the dataset for testing and the other remaining columns the IoU for all classes and the IoU for just the greenery classes (Tree and Vegetation Misc for CamVid). The results show a significant improvement for the CamVid dataset ($80\%$ against $59\%$) and even better for the Greenery (INACITY) dataset ($80\%$ against $35\%$). 
Image~\ref{img:camvid_original} shows one frame from CamVid, Image~\ref{img:camvid_GT} shows its Ground Truth (GT), Images~\ref{img:camvid_no_weighted_pred} and~\ref{img:camvid_weighted_pred} show its prediction for the EDS-32 without and with the weighting strategy, respectively.
For the reduced number of class strategy, we have $86\%$ against $80\%$ for CamVid but for Greeney (INACITY) we have a decrease in performance ($80\%$ against $63\%$). That is, the strategy of using less classes favors the CamVid but not the in-house dataset. As expected, the results show a significant improvement for the IoU greenery in CamVid's dataset.
\begin{table}[ht]
\centering
\caption{EDS mean IoU score for all classes and only for the greenery classes (Tree and VegetationMisc) over CamVid and Greenery datasets.}
\begin{tabular}{|c|c|c|c|}
\hline
N. classes & Dataset & IoU & IoU greenery \\ \hline
7 & Greenery (INACITY)  & 63\% & 63\% \\
7 & CamVid              & 86\% & 65\% \\
32 & Greenery (INACITY) & 80\% & 80\% \\
32 & CamVid             & 80\% & 61\% \\ \hline
\end{tabular}
\label{tab:enc_camvid_greenery}
\end{table}

Table~\ref{tab:comparison_treepedia2} shows the comparison between the EDS trained with 7 classes (EDS-7) and the results reported in~\cite{cai2018treepedia} (Semantic Segmentation Network - SSN - and the Regression Retwork - RN - referred as DCNN end-to-end). We adopted the same metrics used in \cite{cai2018treepedia} to establish the comparison, that is, greenery's IoU (IoU g.), Mean Absolute Error (MAE), Pearson's Correlation Coeficient (PCC) and Estimation Error for GVI in the 5\%-95\% confidence range.

\begin{table}[ht]
\centering
\caption{Comparison between EDS-7 and results reported in~\cite{cai2018treepedia}.}
\begin{tabular}{|c|c|c|c|c|}
\hline
Architecture   & IoU g. & MAE (\%) & PCC   & EE (\%) \\ \hline
EDS-7         &    65.5\%    & 4.12     & 0.865 & -5.4, 26.4 \\ \hline
SSN  &    61.3\%    & 7.83     & 0.830 & -20, 12.3 \\ \hline
RN &    NA        & 4.67     & 0.939 & -10, 7.9 \\ \hline
\end{tabular}
\label{tab:comparison_treepedia2}
\end{table}

\section{Discussion and conclusion}
\label{sec:discussion_conclusion}

In this paper we propose a weighting strategy to improve results obtained by means 
of DLN. Using the aforementioned strategy we could improve previous MSE and IoU results obtained by Li et al.~\cite{cai2018treepedia} with respect to their semantic segmentation network (adapted PSPNet). Our approach using the weighted LF achieved $65.5\%$ in the CamVid's dataset greenery's IoU with the EDS architecture trained over CamVid with 7 classes, an improvement of nearly 4\%. 
Nevertheless their end-to-end Regression Network showed better results with respect to Pearson's Correlation Coefficient and an Estimation Error (EE) less spread apart.

We compared our approach with different architectures in order to demonstrate the advantages obtained by using a per-image weighting strategy. 

Finally, we noticed that reducing the number of labels resulted in a poorer performance for the in-house dataset but not for the CamVid dataset. The network trained with 32 classes produced results with a recall rate reasonably greater than the one with 7 classes. On the other hand the latter produced more precise results, but with a smaller margin over the former's precision when compared to the their respective recall rate. We plan to investigate this labeling reducing strategy better in the future.

\bibliographystyle{IEEEbib}
\bibliography{strings,refs,bibliografia}

\begin{thebibliography}{10}

\bibitem{lovasi2008children}
Gina~Schellenbaum Lovasi, James~W Quinn, Kathryn~M Neckerman, Matthew~S
  Perzanowski, and Andrew Rundle,
\newblock ``Children living in areas with more street trees have lower
  prevalence of asthma,''
\newblock {\em Journal of Epidemiology \& Community Health}, vol. 62, no. 7,
  pp. 647--649, 2008.

\bibitem{mcpherson1997quantifying}
E~Gregory McPherson, David Nowak, Gordon Heisler, Sue Grimmond, Catherine
  Souch, Rich Grant, and Rowan Rowntree,
\newblock ``Quantifying urban forest structure, function, and value: the
  chicago urban forest climate project,''
\newblock {\em Urban ecosystems}, vol. 1, no. 1, pp. 49--61, 1997.

\bibitem{nowak2014tree}
David~J Nowak, Satoshi Hirabayashi, Allison Bodine, and Eric Greenfield,
\newblock ``Tree and forest effects on air quality and human health in the
  united states,''
\newblock {\em Environmental Pollution}, vol. 193, pp. 119--129, 2014.

\bibitem{kardan2015neighborhood}
Omid Kardan, Peter Gozdyra, Bratislav Misic, Faisal Moola, Lyle~J Palmer,
  Tom{\'a}{\v{s}} Paus, and Marc~G Berman,
\newblock ``Neighborhood greenspace and health in a large urban center,''
\newblock {\em Scientific Reports}, vol. 5, pp. 11610, 2015.

\bibitem{lothian1999landscape}
Andrew Lothian,
\newblock ``Landscape and the philosophy of aesthetics: is landscape quality
  inherent in the landscape or in the eye of the beholder?,''
\newblock {\em Landscape and urban planning}, vol. 44, no. 4, pp. 177--198,
  1999.

\bibitem{thayer1978plants}
Robert~L Thayer and Brian~G Atwood,
\newblock ``Plants, complexity, and pleasure in urban and suburban
  environments,''
\newblock {\em Journal of Nonverbal Behavior}, vol. 3, no. 2, pp. 67--76, 1978.

\bibitem{yang2009can}
Jun Yang, Linsen Zhao, Joe Mcbride, and Peng Gong,
\newblock ``Can you see green? assessing the visibility of urban forests in
  cities,''
\newblock {\em Landscape and Urban Planning}, vol. 91, no. 2, pp. 97--104,
  2009.

\bibitem{li2015assessing}
Xiaojiang Li, Chuanrong Zhang, Weidong Li, Robert Ricard, Qingyan Meng, and
  Weixing Zhang,
\newblock ``Assessing street-level urban greenery using google street view and
  a modified green view index,''
\newblock {\em Urban Forestry \& Urban Greening}, vol. 14, no. 3, pp. 675--685,
  2015.

\bibitem{cai2018treepedia}
Bill~Yang Cai, Xiaojiang Li, Ian Seiferling, and Carlo Ratti,
\newblock ``Treepedia 2.0: Applying deep learning for large-scale
  quantification of urban tree cover,''
\newblock {\em arXiv preprint arXiv:1808.04754}, 2018.

\bibitem{DBLP:journals/corr/ZhaoSQWJ16}
Hengshuang Zhao, Jianping Shi, Xiaojuan Qi, Xiaogang Wang, and Jiaya Jia,
\newblock ``Pyramid scene parsing network,''
\newblock {\em CoRR}, vol. abs/1612.01105, 2016.

\bibitem{Cordts2016Cityscapes}
Marius Cordts, Mohamed Omran, Sebastian Ramos, Timo Rehfeld, Markus Enzweiler,
  Rodrigo Benenson, Uwe Franke, Stefan Roth, and Bernt Schiele,
\newblock ``The cityscapes dataset for semantic urban scene understanding,''
\newblock in {\em Proc. of the IEEE Conference on Computer Vision and Pattern
  Recognition (CVPR)}, 2016.

\bibitem{DBLP:journals/corr/HeZRS15}
Kaiming He, Xiangyu Zhang, Shaoqing Ren, and Jian Sun,
\newblock ``Deep residual learning for image recognition,''
\newblock {\em CoRR}, vol. abs/1512.03385, 2015.

\bibitem{DBLP:journals/corr/PohlenHML16}
Tobias Pohlen, Alexander Hermans, Markus Mathias, and Bastian Leibe,
\newblock ``Full-resolution residual networks for semantic segmentation in
  street scenes,''
\newblock {\em CoRR}, vol. abs/1611.08323, 2016.

\bibitem{DBLP:journals/corr/ChenPK0Y16}
Liang{-}Chieh Chen, George Papandreou, Iasonas Kokkinos, Kevin Murphy, and
  Alan~L. Yuille,
\newblock ``Deeplab: Semantic image segmentation with deep convolutional nets,
  atrous convolution, and fully connected crfs,''
\newblock {\em CoRR}, vol. abs/1606.00915, 2016.

\bibitem{DBLP:journals/corr/LinMS016}
Guosheng Lin, Anton Milan, Chunhua Shen, and Ian~D. Reid,
\newblock ``Refinenet: Multi-path refinement networks for high-resolution
  semantic segmentation,''
\newblock {\em CoRR}, vol. abs/1611.06612, 2016.

\bibitem{Yang_2018_CVPR}
Maoke Yang, Kun Yu, Chi Zhang, Zhiwei Li, and Kuiyuan Yang,
\newblock ``Denseaspp for semantic segmentation in street scenes,''
\newblock in {\em The IEEE Conference on Computer Vision and Pattern
  Recognition (CVPR)}, June 2018.

\bibitem{DBLP:journals/corr/abs-1802-02611}
Liang{-}Chieh Chen, Yukun Zhu, George Papandreou, Florian Schroff, and Hartwig
  Adam,
\newblock ``Encoder-decoder with atrous separable convolution for semantic
  image segmentation,''
\newblock {\em CoRR}, vol. abs/1802.02611, 2018.

\bibitem{valada2017adapnet}
Abhinav Valada, Johan Vertens, Ankit Dhall, and Wolfram Burgard,
\newblock ``Adapnet: Adaptive semantic segmentation in adverse environmental
  conditions,''
\newblock in {\em Robotics and Automation (ICRA), 2017 IEEE International
  Conference on}. IEEE, 2017, pp. 4644--4651.

\bibitem{fauqueur2007assisted}
Julien Fauqueur, Gabriel Brostow, and Roberto Cipolla,
\newblock ``Assisted video object labeling by joint tracking of regions and
  keypoints,''
\newblock in {\em Computer Vision, 2007. ICCV 2007. IEEE 11th International
  Conference on}. IEEE, 2007, pp. 1--7.

\bibitem{github:groundtruth}
Oliveira A. A.~A. M.,
\newblock ``Icip 2019 greenery deeplearning github repository,''
  \url{https://github.com/arturandre/ICIP2019_greenery_deeplearning}, 2019,
\newblock Last access in 01/2019.

\bibitem{badrinarayanan2015segnet}
Vijay Badrinarayanan, Alex Kendall, and Roberto Cipolla,
\newblock ``Segnet: A deep convolutional encoder-decoder architecture for image
  segmentation,''
\newblock {\em arXiv preprint arXiv:1511.00561}, 2015.

\bibitem{DBLP:journals/corr/SimonyanZ14a}
Karen Simonyan and Andrew Zisserman,
\newblock ``Very deep convolutional networks for large-scale image
  recognition,''
\newblock {\em CoRR}, vol. abs/1409.1556, 2014.

\bibitem{semsegsuiten}
George Seif,
\newblock ``Semantic segmentation suite in tensorflow,''
  \url{https://github.com/GeorgeSeif/Semantic-Segmentation-Suite}, 2018,
\newblock Last acess 01/2019.

\bibitem{DBLP:journals/corr/RonnebergerFB15}
Olaf Ronneberger, Philipp Fischer, and Thomas Brox,
\newblock ``U-net: Convolutional networks for biomedical image segmentation,''
\newblock {\em CoRR}, vol. abs/1505.04597, 2015.

\bibitem{DBLP:journals/corr/HowardZCKWWAA17}
Andrew~G. Howard, Menglong Zhu, Bo~Chen, Dmitry Kalenichenko, Weijun Wang,
  Tobias Weyand, Marco Andreetto, and Hartwig Adam,
\newblock ``Mobilenets: Efficient convolutional neural networks for mobile
  vision applications,''
\newblock {\em CoRR}, vol. abs/1704.04861, 2017.

\bibitem{treepedia}
MIT Senseable~City Lab,
\newblock ``Treepedia :: Mit senseable city lab,''
  \url{http://senseable.mit.edu/treepedia}, 2017,
\newblock Last access in 01/2019.

\bibitem{harris2010digital}
David Harris and Sarah Harris,
\newblock {\em Digital design and computer architecture},
\newblock Morgan Kaufmann, 2010.

\bibitem{ronneberger2015u}
Olaf Ronneberger, Philipp Fischer, and Thomas Brox,
\newblock ``U-net: Convolutional networks for biomedical image segmentation,''
\newblock in {\em International Conference on Medical image computing and
  computer-assisted intervention}. Springer, 2015, pp. 234--241.

\end{thebibliography}

\end{document}